\documentclass[11pt]{article}

\usepackage[a4paper,margin=1in]{geometry}
\usepackage{times}
\usepackage{graphicx}
\usepackage{xcolor}
\usepackage{amsmath,amssymb}
\usepackage{booktabs}
\usepackage{enumitem}
\usepackage{caption}
\usepackage{subcaption}
\usepackage{hyperref}
\usepackage{titlesec}
\usepackage[most]{tcolorbox}
\usepackage[numbers,sort&compress]{natbib}
\usepackage{multirow} 
\usepackage{subcaption}

\definecolor{seedblue}{RGB}{49,91,157}
\definecolor{lightblue}{RGB}{235,242,252}

\hypersetup{
    colorlinks=true,
    linkcolor=seedblue,
    citecolor=seedblue,
    urlcolor=seedblue
}

\titleformat{\section}
  {\Large\bfseries\color{seedblue}}
  {\thesection}{0.8em}{}

\titleformat{\subsection}
  {\large\bfseries\color{seedblue}}
  {\thesubsection}{0.8em}{}

\titleformat{\subsubsection}
  {\normalsize\bfseries\color{seedblue}}
  {\thesubsubsection}{0.8em}{}

\definecolor{color1}{rgb}{1.        , 0.62352941, 0.16862745}
\definecolor{color2}{rgb}{0.11764706, 0.27843137, 0.68235294}
\definecolor{color3}{rgb}{0.8627451 , 0.08627451, 0.29019608}
\definecolor{color4}{rgb}{0.31372549, 0.70196078, 0.1372549}
\definecolor{color5}{rgb}{0.46666667, 0.13333333, 0.02352941}

\def\ourname{PhysX-Omni}
\def\ournewdata{PhysXVerse}
\def\ourbench{PhysX-Bench}

\newcommand{\reporttitle}{
PhysX-Omni: Unified Simulation-Ready Physical 3D Generation for Rigid, Deformable, and Articulated Objects
}

\newcommand{\reportauthor}{
Ziang Cao$^1$, Yinghao Liu$^2$, Haitian Li$^1$, Runmao Yao$^1$,  Fangzhou Hong$^1$, \\ Zhaoxi Chen$^1$, Liang Pan$^2$, Ziwei Liu$^1$
}

\newcommand{\officialpage}{
\href{https://physx-omni.github.io/}{https://physx-omni.github.io/}
}

\begin{document}


\noindent
\raisebox{0pt}{\includegraphics[height=0.7in]{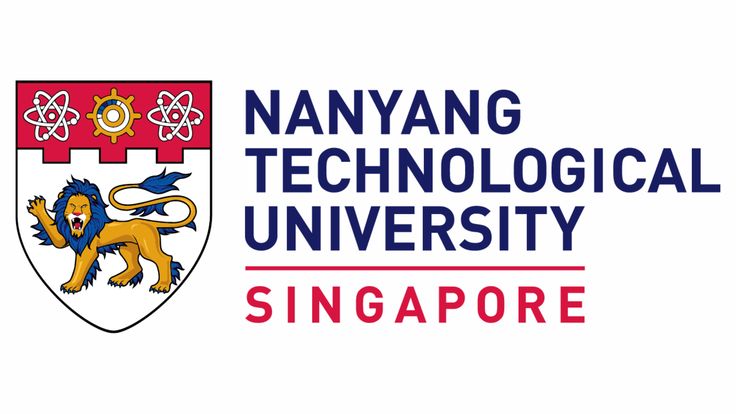}}
\hspace{-0.3em}
\raisebox{4pt}{\includegraphics[height=0.5in]{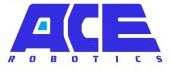}}

\vspace{-20pt}
\begin{center} 
    \rule{\linewidth}{0.5pt}

    \vspace{1.2em}
    {\LARGE\bfseries \reporttitle}

    \vspace{0.5em}
    \rule{\linewidth}{0.5pt}

    \vspace{1em}
    {\large\bfseries \reportauthor}

    \vspace{0.5em}
    $^1$ S-Lab, Nanyang Technological University, $^2$ ACE Robotics
\end{center}

\begin{center}
    \includegraphics[width=1\textwidth]{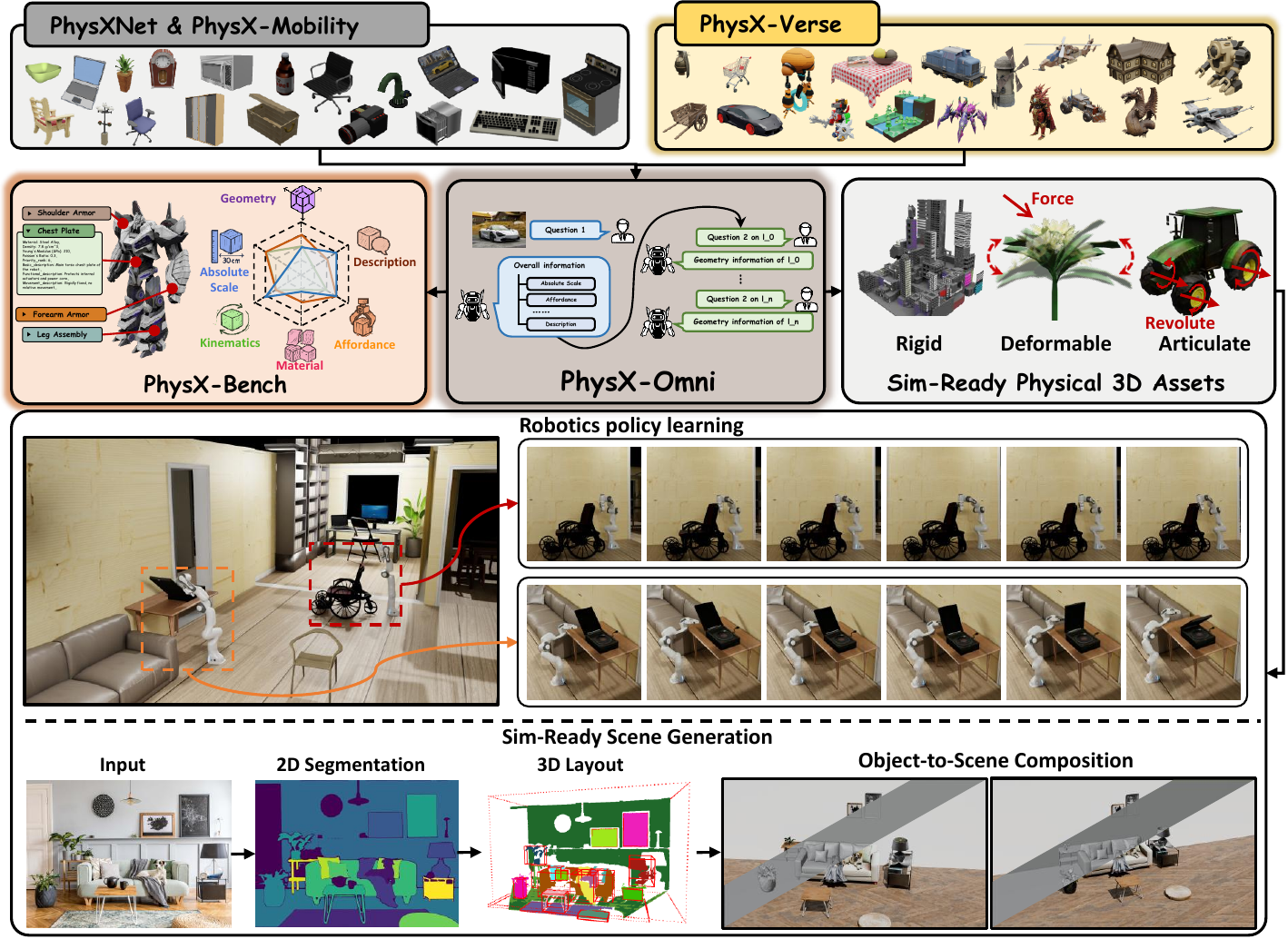}
    
    \captionof{figure}{
    By exploiting the high diversity of \ournewdata, \ourname\ is capable of generating detailed and general 3D assets covering rigid, deformable, and articulated objects, producing simulation-ready physical assets suitable for downstream applications.
    }
    \label{fig:teaser}
\end{center}

\begin{tcolorbox}[
    breakable,
    enhanced,
    colback=white,
    colframe=seedblue,
    arc=3pt,
    boxrule=0.6pt,
    left=12pt,
    right=12pt,
    top=10pt,
    bottom=10pt
]
\begin{center}
    {\large\bfseries\color{seedblue} Abstract}
\end{center}

Simulation-ready physical 3D assets have emerged as a promising direction owing to their broad applicability in downstream tasks. However, most existing 3D generation methods either neglect physical properties or are limited to a single asset category, \textit{e.g.}, rigid, deformable, or articulated objects. To address these limitations, we introduce \textbf{\ourname}, \textbf{a unified framework} for simulation-ready physical 3D generation across diverse asset types.
Specifically, we develop a novel and efficient geometry representation tailored for Vision-Language Models, which directly encodes high-resolution 3D structures without compression, significantly improving generation performance. In addition, we construct the \textbf{first general simulation-ready 3D dataset}, \textbf{\ournewdata}, covering diverse indoor and outdoor categories. Furthermore, to comprehensively and flexibly evaluate both generative and understanding capabilities in the wild, we propose \textbf{\ourbench}, which encompasses six key attributes: geometry, \textbf{\textcolor{color2}{absolute scale}}, \textbf{\textcolor{color3}{material}}, \textbf{\textcolor{color1}{affordance}}, \textbf{\textcolor{color4}{kinematics}}, and \textbf{\textcolor{color5}{description}}.
Extensive experiments with conventional metrics and \ourbench\ show that \ourname\ performs strongly in both generation and understanding. Moreover, additional studies further validate the potential of \ourname\ for applications in simulation-ready scene generation and robotic policy learning. We believe \ourname\ can significantly advance a wide range of downstream applications, particularly in embodied AI and physics-based simulation.

\vspace{0.5em}
\noindent\textbf{Official Page:} \officialpage

\noindent\textbf{Correspondence:} Ziwei Liu (\href{mailto:author@email.com}{ziwei.liu@ntu.edu.sg})
\end{tcolorbox}

\vspace{1em}


\clearpage

\tableofcontents
\clearpage

\section{Introduction}

High-quality simulation-ready (sim-ready) 3D assets have attracted significant attention due to their wide range of downstream applications in gaming design, robotics, embodied AI, and interactive simulation. However, most existing 3D generation approaches primarily focus on achieving photorealistic appearance and detailed geometric structures~\cite{trellis,3dtopia,3dtopiaxl,fang2025meshllm,ye2025shapellm,trellis2,zhang2025bang,yang2025omnipart}. Despite their strong generative performance, the generated 3D assets often lack essential physical attributes required for real-world deployment, thereby limiting their applicability, particularly in physics-based scenarios.

To bridge this gap, a number of works have focused on generating articulated assets~\cite{chen2024urdformer,liu2024singapo,le2024articulate,lu2025dreamart,li2026monoart} and deformable assets~\cite{zhang2024physdreamer,guo2024physically,chen2025physgen3d,le2025pixie,chen2025vid2sim,jiang2025phystwin}. However, these methods typically model only a limited subset of physical attributes for a specific asset type (\textit{e.g.}, articulated or deformable objects), while overlooking other essential properties. As pioneering efforts in sim-ready physical 3D generation~\cite{cao2025physx,cao2025physxanything}, they enable the synthesis of richer physical attributes. Nevertheless, they remain constrained by the scarcity of large-scale, high-quality annotated 3D datasets, which limits the diversity of generated assets and, consequently, their practical utility for downstream embodied AI and control tasks. Furthermore, the absence of effective benchmarks for evaluating physical attributes in real-world scenarios (without ground-truth annotations) significantly limits meaningful evaluation.

To address these challenges, we propose \textbf{\ourname}, a unified simulation-ready physical 3D generative framework that supports diverse object types, including rigid, deformable, and articulated assets , with broad potential applications as illustrated in Fig.~\ref{fig:teaser}. Specifically, we introduce a novel geometry representation tailored for Vision-Language Models (VLM), which directly models high-resolution 3D structures without requiring additional special tokens during training. By explicitly modeling 3D structure, \ourname\ avoid the failure modes caused by segmentation, thereby significantly improving generative performance. Moreover, since we avoid additional decoder refinement, our framework remains compatible with existing voxel-based decoders~\cite{trellis,trellis2,ren2024xcube}, enabling the synthesis of high-fidelity appearance. 

To address data scarcity, we construct the first general simulation-ready physical 3D dataset, \textbf{\ournewdata}, which contains over 8K assets spanning more than 2K indoor and outdoor categories, \textit{e.g.}, helicopters, tanks, racing cars, skyscrapers, and toys, curated and filtered from PartVerse~\cite{dong2025one}. Furthermore, to comprehensively evaluate simulation-ready 3D generation, we build the first physical 3D generative benchmark, \textbf{\ourbench}, covering six key attributes: geometry, \textbf{\textcolor{color2}{absolute scale}}, \textbf{\textcolor{color3}{material}}, \textbf{\textcolor{color1}{affordance}}, \textbf{\textcolor{color4}{kinematics}}, and \textbf{\textcolor{color5}{description}}. By leveraging physics-based simulation and powerful VLMs, \ourbench\ enables robust and realistic evaluation in in-the-wild scenarios. Comprehensive experiments with conventional metrics and \ourbench\ demonstrate that \ourname\ achieves superior performance in both generation quality and generalization compared to recent state-of-the-art methods. Finally, to validate deployability in standard simulators and physics engines, we conduct experiments in a common simulation environment, showing that our simulation-ready assets can be directly applied to contact-rich robotic policy learning. We believe our work opens up new opportunities for future research in 3D generation, embodied AI, and robotics.

To summarize, our main contributions are: 
\begin{itemize}

\item We introduce \textbf{\ourname}, a novel unified framework for simulation-ready physical 3D generation across diverse asset types. By employing the new tailored geometry representation, our approach directly models detailed geometric structures, leading to significantly improvements in both performance and generalization.

    \item We construct the first general simulation-ready physical 3D dataset, \textbf{\ournewdata}, covering over 2K indoor and outdoor categories (\textit{e.g.}, trucks, jets, and flowers), with high-quality physical attribute annotations.

    \item We introduce the first benchmark for simulation-ready physical 3D generation, \textbf{\ourbench}. By integrating physics-based simulation with powerful VLMs, \ourbench\ provides a comprehensive and robust evaluation framework for assessing generation methods in real-world scenarios across six key attributes.
   
    \item Extensive evaluations on \ourbench\ and conventional benchmarks demonstrate that \ourname\ achieves impressive generative quality and robust generalization. Moreover, we verify the deployability of our simulation-ready assets in standard simulation environments, facilitating downstream applications in embodied AI and robotic manipulation.

\end{itemize}

\begin{figure*}[t]
		\centering	
		
        \includegraphics[width=1\linewidth]{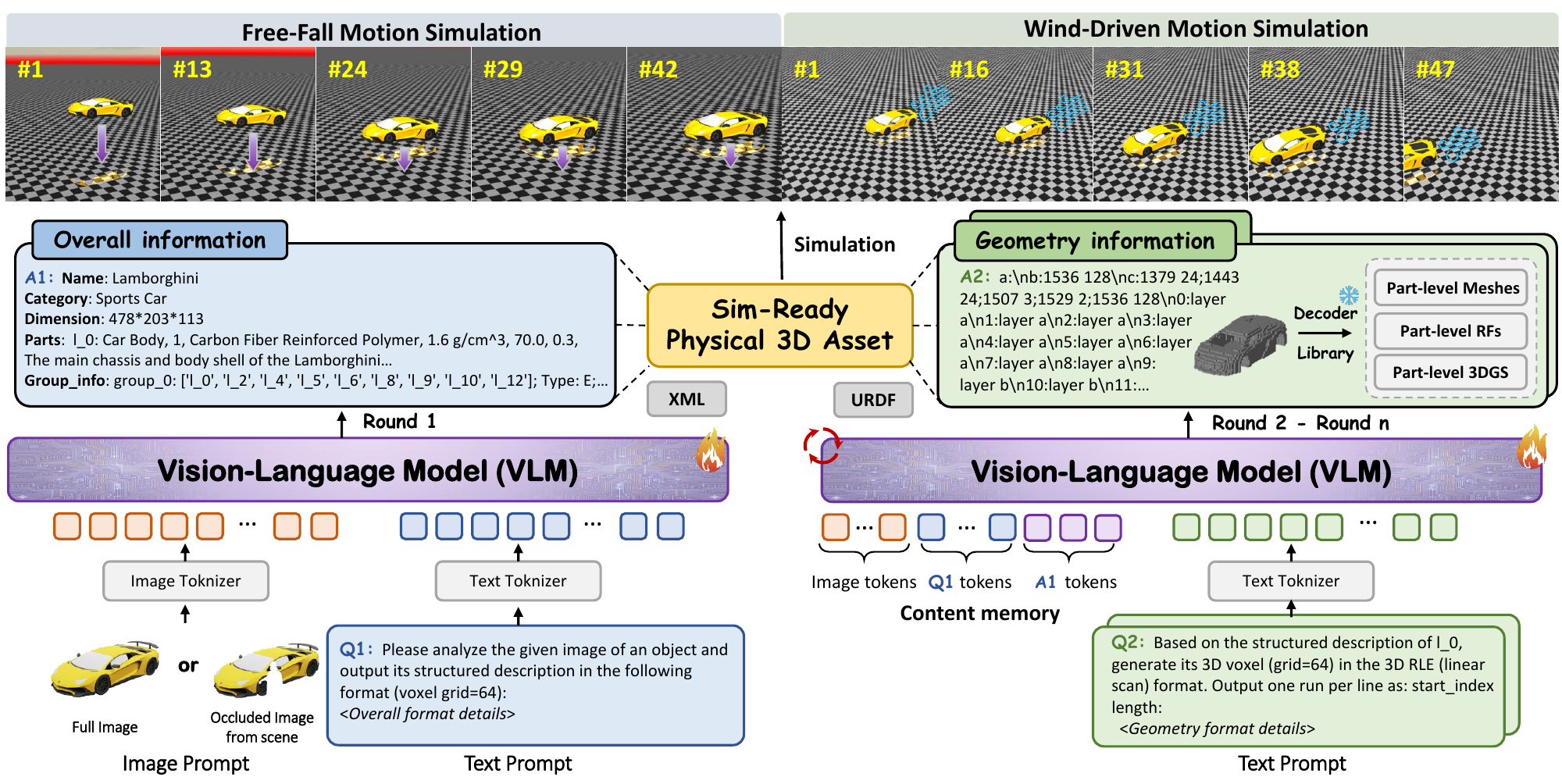}
		\vspace{-6mm}
		\caption{Given a single complete or partially occluded image, \ourname\ first infers high-level overall information. It then employs a multi-turn generation process to produce detailed part-level geometry. Owing to the inherent alignment between global and local representations, these outputs can be directly integrated into simulation-ready physical 3D assets.}

		\label{fig:framework}
		
	\end{figure*}

\section{Related Works}
\label{sec:related}

    




\subsection{Appearance-Centric 3D Generation}

Early efforts in 3D generation were largely dominated by generative adversarial networks (GANs), which laid the foundation for this field~\cite{chan2022efficient,gao2022get3d}. Despite their initial success, GAN-based approaches often suffer from instability and limited robustness when scaling to more complex and diverse data distributions. The introduction of DreamFusion~\cite{poole2022dreamfusion} marked a significant shift by proposing score distillation sampling (SDS), which leverages the strong priors of pretrained 2D diffusion models. Nevertheless, such optimization-based pipelines remain computationally expensive and are prone to artifacts such as the Janus effect. To address these limitations, recent works increasingly favor feed-forward architectures, which offer improved efficiency and more stable generation behavior~\cite{trellis,lgm,instantmesh,3dtopia,3dtopiaxl,difftf,difftf1,zlabor,yang2025holopart,yao2026anchoreddream,lin2025partcrafter,seed2025seed3d}. In parallel, alternative paradigms have also been explored, including autoregressive approaches that model 3D structures sequentially~\cite{chen2024meshanything,siddiqui2024meshgpt}. To mitigate the challenge of long token sequences in geometry modeling, LLaMA-Mesh~\cite{wang2024llama} adopts a simplified mesh representation, while MeshLLM~\cite{fang2025meshllm} introduces a hierarchical part-level generation strategy to further improve quality. ShapeLLM-Omni~\cite{ye2025shapellm} instead compresses 3D representations via a VQ-VAE, but at the cost of introducing specialized tokens and a dedicated tokenizer, which complicates the training pipeline.

In contrast, PhysX-Anything~\cite{cao2025physxanything} explores modeling simulation-ready physical 3D assets using pure text representations. Benefiting from the strong prior knowledge of VLMs, it achieves impressive generative performance and robustness. However, its reliance on an explicit segmentation stage introduces a performance bottleneck, as the overall quality is constrained by the segmentation module. To overcome this limitation, we propose a new geometry representation that directly models high-resolution 3D structures. By simplifying the overall framework, our approach significantly improves generation performance over the baseline.

\subsection{Physical 3D Asset Generation}

Articulated object generation has recently gained increasing attention due to its broad range of downstream applications~\cite{li2026monoart,qiu2025articulate,song2025magicarticulate,su2025artformer,chen2025artilatent,chen2025freeart3d,joshi2505procedural}. Existing articulate generation approaches can be broadly categorized into several paradigms. A dominant line of work follows a retrieval-based strategy, where articulated assets are constructed by retrieving and assembling meshes from a predefined source library~\cite{chen2024urdformer,le2024articulate}. While effective within known categories, such methods are inherently limited by the coverage of the database and struggle to generalize to novel structures. Another line of research adopts graph-structured representations~\cite{liu2024singapo,lei2023nap}, integrating kinematic graphs with diffusion models to enable structure-aware generation. However, these approaches typically focus on geometry and lack the ability to produce high-quality textured assets, limiting their realism. Beyond these paradigms, optimization-based methods such as DreamArt~\cite{lu2025dreamart} attempt to reconstruct articulated objects from video generation outputs. Despite their flexibility, they rely on manually annotated part masks and tend to become unstable when handling objects with many movable components. URDF-Anything~\cite{li2025urdf} and URDF-Anything+~\cite{wu2026urdf} directly generates URDF representations, but its performance heavily depends on high-quality point cloud inputs or mesh and it remains challenging to produce detailed textures. Recently, MonoArt~\cite{li2026monoart} leverages priors from 3D generation and segmentation to infer kinematic parameters and achieve promising performance. Nevertheless, all those method primarily focuses on a single type of physical attribute and lacks a holistic modeling of physical objects. Beyond articulated object generation, several works have also explored modeling the deformation of 3D assets~\cite{chen2025physgen3d,le2025pixie,chen2025vid2sim,jiang2025phystwin,cao2025sophy}. However, these approaches also overlook other critical physical attributes, limiting their realism. To advance 3D generation toward physical fidelity, PhysXGen~\cite{cao2025physx} introduces a unified framework that directly generates 3D assets with essential physical properties, such as absolute scale and density. Building upon this line of work, PhysX-Anything~\cite{cao2025physxanything} further extends the paradigm to simulation-ready 3D asset generation. Nevertheless, it remains constrained by the limited diversity of available simulation-ready datasets and faces challenges in modeling high-quality, detailed assets efficiently.

To address these limitations, we propose a tailored geometry representation within a unified framework, along with the first general high-quality simulation-ready 3D dataset. Benefiting from both the enriched data diversity and the efficient geometry representation, our \ourname\ demonstrates strong robustness and superior performance in generating complex topologies and accurate physical attributes. We believe our approach opens up a promising direction for leveraging synthetic data to advance downstream applications.

\begin{figure}[t]
    \centering

    \begin{subfigure}[t]{0.59\linewidth}
        \centering
        \includegraphics[width=\linewidth]{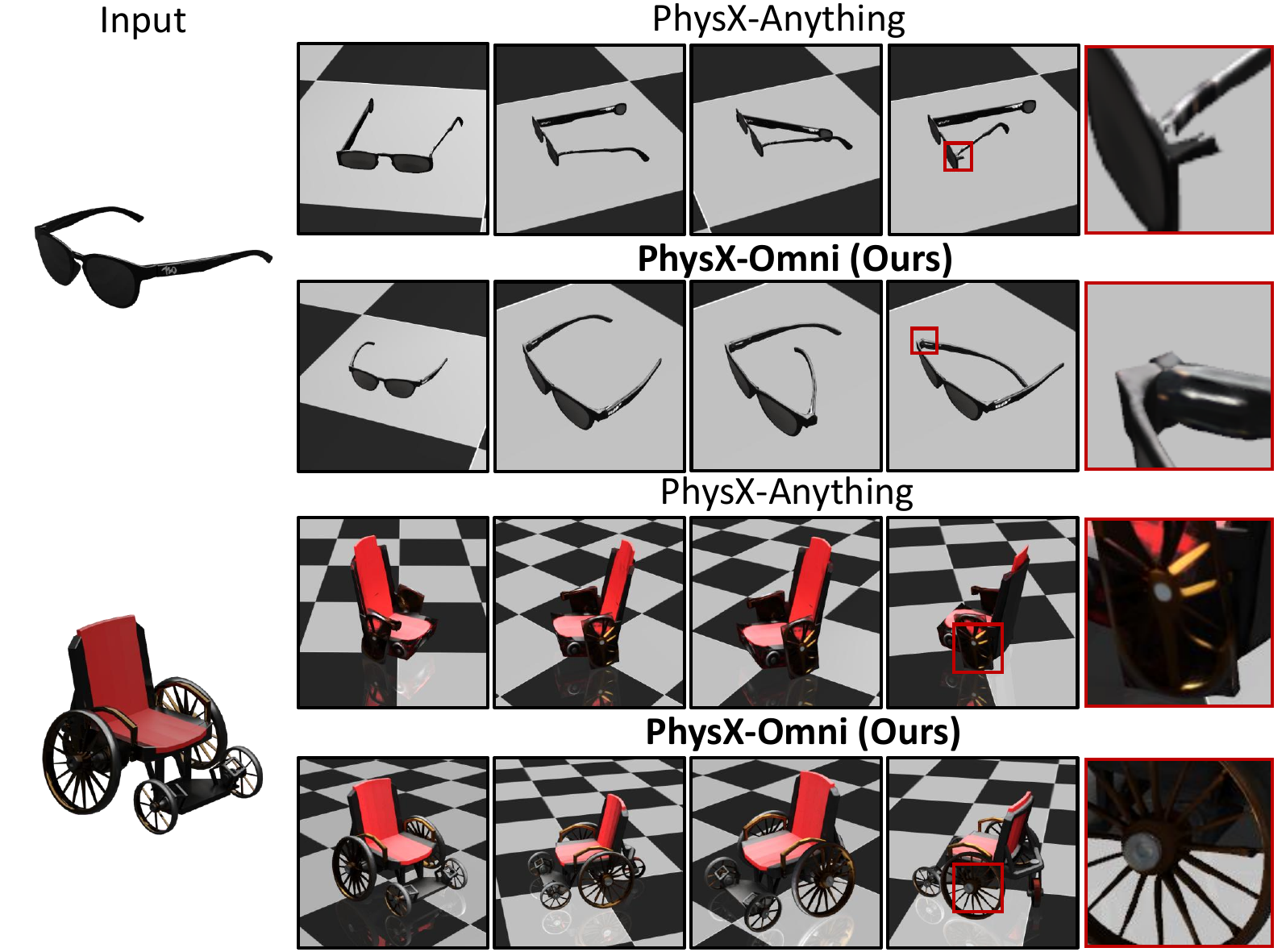}
        \caption{}
        \label{fig:geocomp}
    \end{subfigure}
    \hfill
    \begin{subfigure}[t]{0.4\linewidth}
        \centering
        \includegraphics[width=\linewidth]{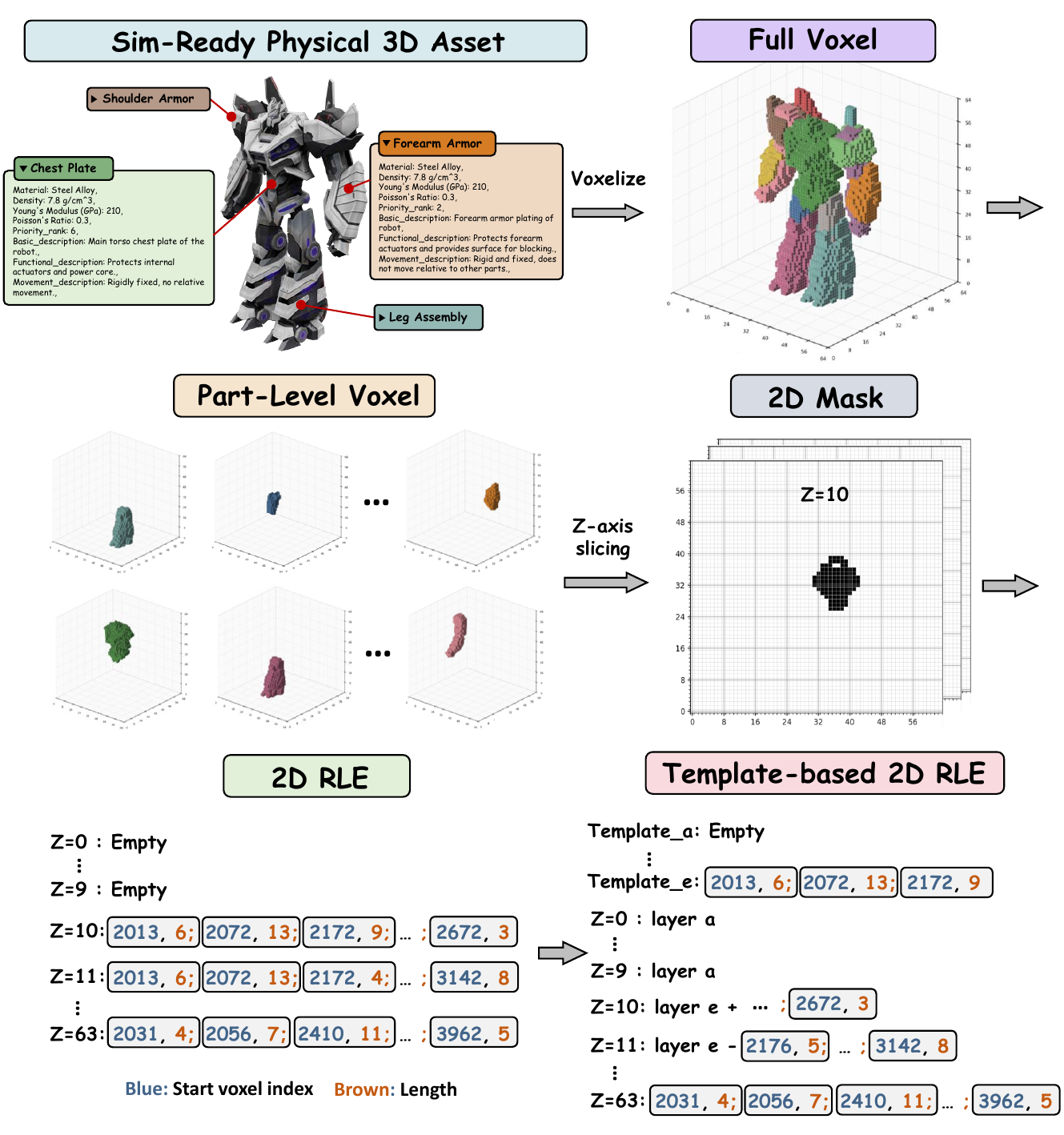}
        \caption{}
        \label{fig:details}
    \end{subfigure}

    \vspace{-2mm}

    \caption{
    \textbf{(a). Comparison of different geometry representations for 3D modeling.} Leveraging the proposed geometry representation, \ourname\ effectively captures fine-grained 3D structures and enhances kinematic accuracy. (b). \textbf{Detailed geometry representation of our \ourname.} To directly model high-resolution 3D structures, we first slice part-level voxel grids along the z-axis. For each resulting 2D mask, we apply classical run-length encoding (RLE) to convert the binary image into a compact textual representation. To further improve compression efficiency, we introduce template layers, enabling other layers to be expressed as variations relative to templates.
    }
    \label{fig:representation}
\end{figure}

\begin{figure*}[t]
		\centering	
		
        \includegraphics[width=1\linewidth]{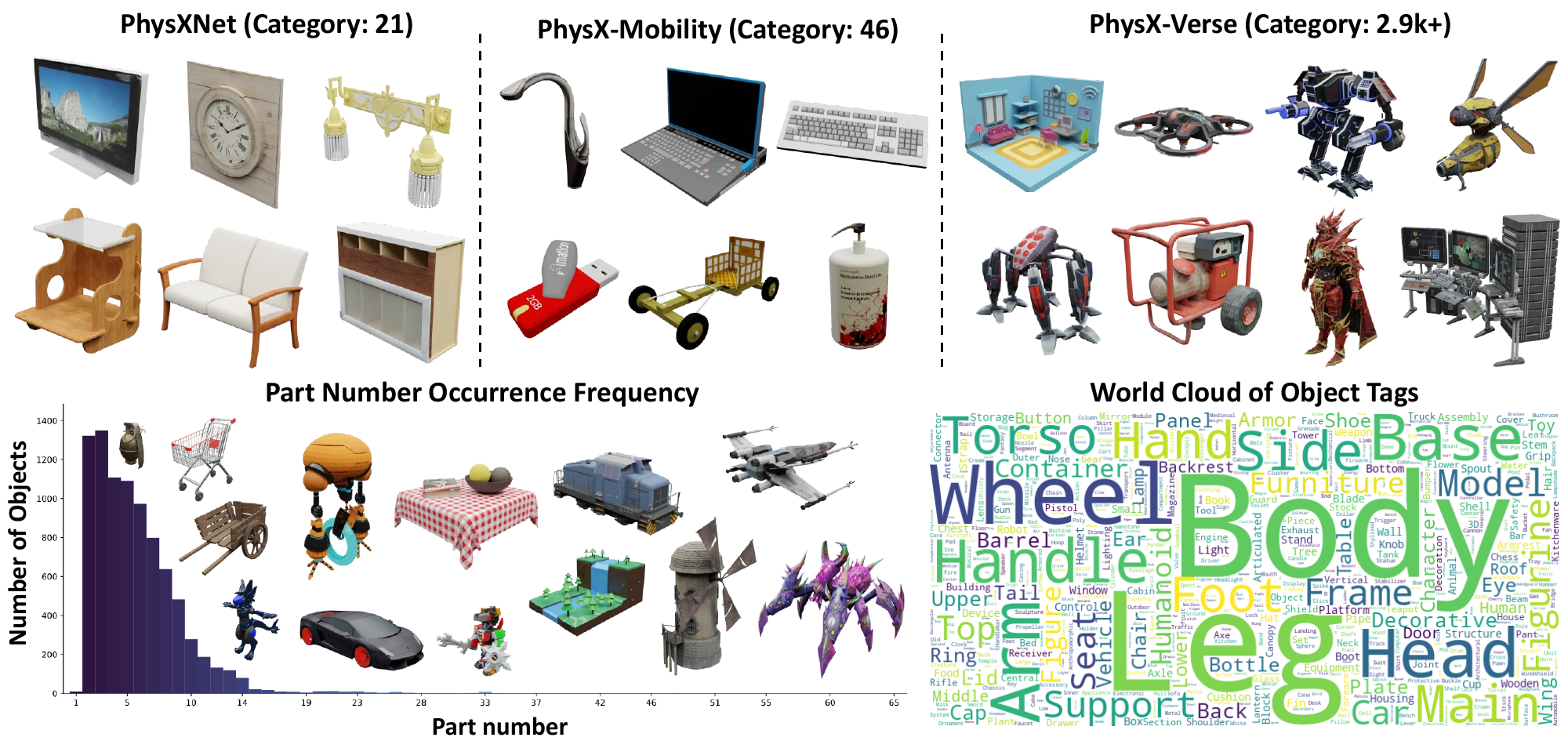}
		\vspace{-6mm}
		\caption{\textbf{Statistics and distribution of \ournewdata.} Compared to existing simulation-ready physical datasets, \ournewdata\ exhibits substantially broader category coverage, including cars, buildings, human models, toys, and robots. The distribution of part counts follows a long-tailed pattern, and the word cloud further illustrates the diversity of semantic categories.}
		\vspace{-6pt}
		\label{fig:datadis}
		
	\end{figure*}

\section{Methodology}

In this section, we describe the core components of \ourname, including the overall paradigm illustrated in Fig.~\ref{fig:framework}, the newly constructed dataset, \ournewdata, and the first benchmark for simulation-ready 3D assets, \ourbench.


\subsection{Generative paradigm of \ourname}

\ourname\ adopts a VLM-based generation paradigm to produce simulation-ready physical assets through a coarse-to-fine global-to-local reasoning process, following~\cite{cao2025physxanything}. As illustrated in Fig.~\ref{fig:framework}, given a complete or partially occluded image, \ourname\ first performs holistic understanding to infer high-level global information, including the object category, semantic identity, absolute scale, component hierarchy, and potential physical properties. Such global understanding provides strong structural and semantic priors for subsequent part-level generation and helps maintain consistency between the overall object and its local components.

Based on the inferred global representation, \ourname\ further predicts the detailed geometric structure and physical attributes of each individual part. For the global representation, we follow the tree-structured and VLM-friendly formulation introduced in~\cite{cao2025physx}, which effectively organizes object-level and part-level information into a hierarchical representation compatible with autoregressive vision--language modeling.

For geometry representation, we introduce a novel high-resolution structure modeling strategy that directly encodes detailed 3D geometry in a compact and generation-friendly manner shown in Fig.~\ref{fig:details}. Unlike prior methods that heavily rely on mesh decomposition or additional segmentation modules, our representation allows \ourname\ to directly model complex geometric structures while preserving explicit structural information. As a result, \ourname\ can seamlessly leverage a pre-trained voxel-based 3D decoder to generate high-quality meshes without requiring additional mesh segmentation processes, thereby significantly improving generation quality, robustness, and generalization ability, especially for objects with complex topologies and fine-grained structures.

Prior works have explored various compact 3D representations for vision--language modeling, including vertex quantization~\cite{wang2024llama,fang2025meshllm}, 3D VQ-GAN representations~\cite{ye2025shapellm}, and text-based voxel indices~\cite{cao2025physxanything} to reduce sequence length and improve generation efficiency. However, these methods either rely on additional special tokens, suffer from limited geometric fidelity, or struggle to explicitly model high-resolution structures in a generation-friendly manner. To address these limitations while maintaining compatibility with existing VLM token spaces, we introduce a novel text-based geometry representation that does not require introducing additional special tokens into the language model vocabulary.

Specifically, inspired by classical 2D run-length encoding (RLE), we propose a template-based RLE representation to explicitly and directly model high-resolution 3D geometry. We first voxelize the simulation-ready assets and decompose them into part-level voxels according to the annotated object structure. Each part-level voxel is then sliced along the z-axis into a sequence of 2D binary masks. For each slice, we apply a compact 2D RLE formulation to encode the occupied regions into text tokens efficiently.

Different from standard 2D RLE, however, 3D structures naturally contain strong spatial redundancy across neighboring slices, especially for smooth or repeated geometric regions. To exploit this property, we further introduce the concept of template layers. Instead of independently encoding every slice, our method allows multiple slices to share a common structural template, while only storing their relative variations or residual differences. By reusing structural patterns across layers, our template-based formulation substantially reduces token redundancy and sequence length while preserving detailed geometric information. Moreover, this design maintains explicit geometric structures throughout the generation process, making it more robust to autoregressive prediction errors and more suitable for high-resolution structure modeling.

As a result, our template-based RLE representation achieves significantly stronger compression efficiency and geometric fidelity compared with conventional 2D RLE and existing text-based explicit representations. We further compare our representation with prior methods in Fig.~\ref{fig:geocomp}. The qualitative results demonstrate that, compared with the baseline using text-based voxel indices, \ourname\ produces substantially more detailed geometric structures and achieves better alignment with physical and kinematic attributes. In particular, our representation enables the model to maintain structural consistency in complex articulated objects while preserving fine-grained geometry. Additional quantitative and qualitative comparisons are provided in the experimental section.

\begin{figure}[t]
		\centering	
		
        \includegraphics[width=0.8\linewidth]{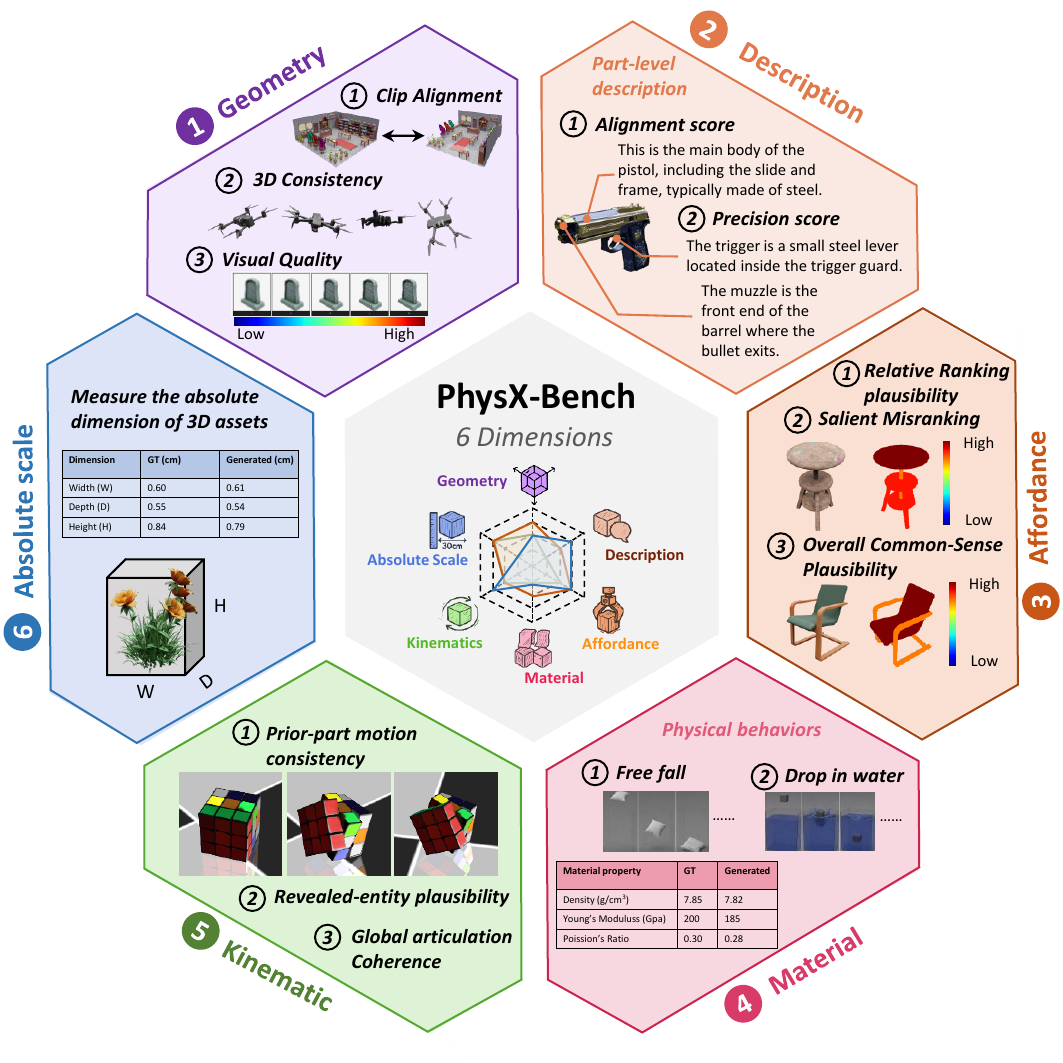}

		\caption{\textbf{Overview of \ourbench.} It consists of six key dimensions for comprehensively evaluating 3D structure, appearance, fundamental physical attributes, and understanding.}
		
		\label{fig:bench}
	\vspace{-10pt}
	\end{figure}


\subsection{\ournewdata\ Datasets}
To alleviate the limitation of data scarcity, we construct the first general simulation-ready physical 3D dataset, \ournewdata. To obtain high-quality simulation-ready assets, we leverage the human-verified segmentation annotations provided by PartVerse~\cite{dong2025one}. For reliable physical properties, we further adopt the human-in-the-loop annotation pipeline introduced in~\cite{cao2025physx}. Specifically, we first preprocess the original dataset by filtering invalid samples and merging excessively small or noisy parts to improve structural consistency. We then render multi-view images of each 3D asset and employ a powerful VLM (GPT) to generate preliminary physical annotations, including absolute scale, affordance, material, functional descriptions, and kinematic information. These automatically generated annotations are subsequently verified and refined by human annotators to ensure both physical plausibility and annotation quality.

As a result, \ournewdata\ contains more than 8.7K high-quality simulation-ready 3D assets spanning over 2.9K categories, covering a wide range of object types, such as indoor furniture, unmanned aerial vehicles, robots, vehicles, and large-scale scene components. Compared with existing simulation-ready datasets, \ournewdata\ exhibits substantially richer category diversity and more comprehensive physical annotations, as illustrated in Fig.~\ref{fig:datadis}. In addition, we analyze the structural complexity of the dataset through the distribution of part counts. The number of parts ranges from 1 to 65, demonstrating that \ournewdata\ covers objects from simple rigid structures to highly complex articulated systems. Such large diversity in both category coverage and structural complexity provides a strong foundation for training and evaluating general simulation-ready physical 3D generation models.

\subsection{Evaluation Dimension of \ourbench}

To guarantee the reproducibility and robustness of the benchmark, we adopt the open-source VLM (Qwen3.5-122B-A10B) to evaluate the generated physical attributes. Moreover, to reduce the difficulty of understanding complex 3D structures and physical properties, we use rendered images or videos as inputs for evaluation rather than directly feeding physical attributes. Our benchmark evaluates six key dimensions: geometry for evaluating 3D structure and appearance, absolute scale for evaluating physical dimensions, affordance for evaluating human–object interaction priors, description for evaluating semantic understanding, material for evaluating mechanical properties, and kinematics for evaluating motion behaviors shown in Fig~\ref{fig:bench}. Specifically, we define three sub-attributes for geometry: \textit{i.e.}, 1) CLIP score to measure the alignment between the generated results and the conditioning image; 2) 3D consistency to assess structural consistency across multi-view renderings; and 3) visual quality to evaluate the appearance quality. To obtain accurate visual quality assessments, we design a reference grading table with five levels ranging from very poor to excellent.

For description evaluation, we render part-level masks on the generated 3D object and use the VLM to evaluate whether the masked regions semantically match the human-annotated reference descriptions from the condition image.
This assesses whether the evaluated generation method preserves and grounds part-level semantics from the condition image in the generated 3D asset.
Since affordance may involve multiple plausible outcomes depending on different functionalities, our evaluation is grounded in human common sense and considers both local and global plausibility, including the relative ranking plausibility and salient misranking of typical parts, as well as the overall rationality of the predicted affordances. Predictions that are more consistent with human common sense will receive higher scores. 
For absolute scale, we compare the maximum generated object dimension with the VLM-estimated maximum real-world dimension and convert the symmetric percentage error into a scale plausibility score.

For the material dimension, we explore evaluating physical properties by rendering the generated assets into different types of simulation videos, mainly including free-fall and water-drop scenarios.
Specifically, the free-fall simulation, particularly the behavior upon ground contact, can reflect properties such as Young’s modulus and Poisson’s ratio; while the water-drop simulation is mainly used to evaluate density.
We believe that evaluating materials through such visualized physical behaviors enables a more intuitive protocol that better aligns with human perception and judgment.
For kinematics, we follow the principle that assets with more reasonable and physically plausible motions should receive higher scores. Specifically, we first render the generated assets into motion videos and then infer potential motions from the conditioning image. For visible parts, we define a \textit{prior-part motion consistency} metric to evaluate whether the predicted motions align with the expected behaviors of observed components. 
For parts that are not visible due to the single-view limitation of the conditioning image but become observable in the rendered motion video, we introduce a \textit{revealed-entity plausibility} metric to assess whether their revealed motions are physically and semantically plausible.
Finally, we define a \textit{global articulation coherence} metric to measure the overall consistency and plausibility of the complete motion dynamics. 
The final kinematics score is computed as a weighted average of the \textit{prior-part motion consistency}, \textit{revealed-entity plausibility}, and \textit{global articulation coherence} scores.

\section{Experiments}

In this section, we present experimental results on both conventional evaluation metrics and our proposed benchmark, \ourbench. In addition, we report the human alignment evaluation results and conduct comprehensive ablation studies to analyze the effectiveness of different components in our framework. Finally, we further demonstrate the potential applications of \ourname\ in downstream simulation-ready scene generation and robotic policy learning tasks.

\subsection{Implementation details}
We adopt Alibaba Cloud Qwen2.5-VL-7B-Instruct as our VLM backbone~\cite{bai2025qwen2}. The model is trained for 5 epochs on 64 NVIDIA A100 GPUs over approximately 14 days, using a peak learning rate of $2\times10^{-5}$, a cosine learning rate decay schedule with a warmup ratio of 0.03, and an effective batch size of 128. To support the generation of high-resolution simulation-ready structures and long-context physical descriptions, we set the maximum sequence length to 16,384 tokens. For the decoding stage, we employ TRELLIS~\cite{trellis} to transform the generated voxel representations into high-quality 3D meshes. Benefiting from our explicit geometry representation, the decoder can directly reconstruct detailed structures without requiring additional mesh segmentation or topology refinement modules, thereby improving both robustness and geometric fidelity. 

\subsection{Datasets}
For training, we combine simulation-ready assets from PhysXNet~\cite{cao2025physx}, PhysX-Mobility~\cite{cao2025physxanything}, and our newly constructed dataset \ournewdata, resulting in a large-scale corpus containing more than 42K simulation-ready physical 3D assets spanning diverse indoor and outdoor categories. The dataset covers rigid, articulated, and deformable objects with rich geometric structures and physical attributes. To improve view consistency and enhance the robustness of visual understanding, we render 25 images for each object from different viewpoints as conditioning inputs during training. This multi-view training strategy enables \ourname\ to better capture the correspondence between visual appearance, geometric structure, and physical properties, leading to stronger generalization performance on complex real-world objects.

\begin{figure*}[t]
		\centering	
		
        \includegraphics[width=1\linewidth]{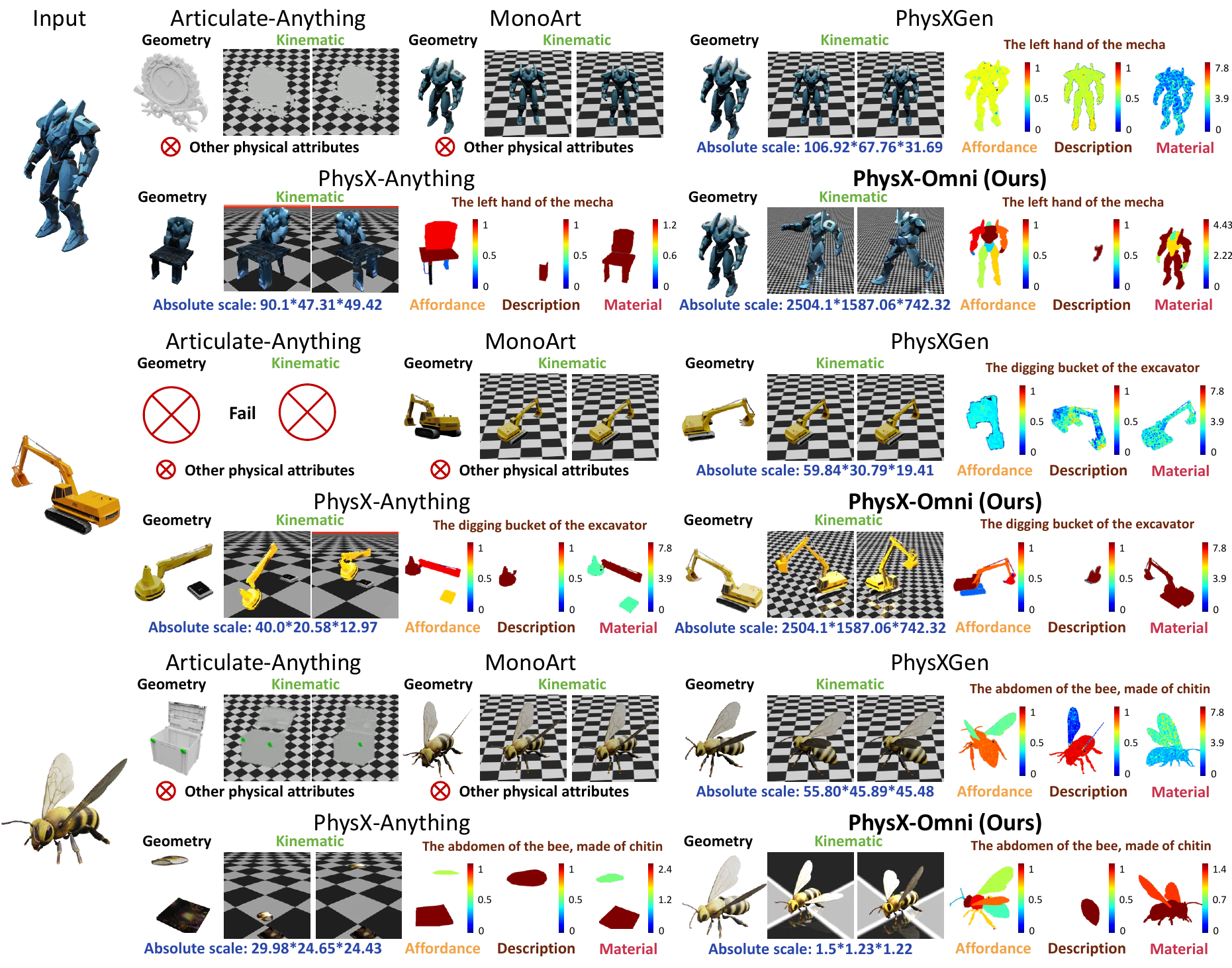}
		\vspace{-8mm}
		\caption{\textbf{Qualitative results.} Compared with existing generative methods, our \ourname\ demonstrates impressive performance in generating complex geometries and rich physical attributes.}
		
		\label{fig:quali}

	\end{figure*}
    \begin{table*}[t]
\caption{\label{table:tradition}\textbf{Quantitative comparison with other methods on conventional metrics.} Note that Chamfer Distance (CD) is reported in units of $\times 10^{-3}$, and F-score (FS) is reported in units of $\times 10^{-2}$ under a distance threshold of 0.05. The results clearly demonstrate the superior generative performance of our method in both geometry and physical attribute generation.}
\vspace{-10pt}
\scriptsize
\centering
\resizebox{1\textwidth}{!}{%

\begin{tabular}{@{}c|c|ccc|ccccc|@{}}
\toprule
\multirow{2}{*}{\bf Dataset} & \multirow{2}{*}{\bf Methods} 
& \multicolumn{3}{c|}{\bf Geometry} 
& \multicolumn{5}{c|}{\bf Physical Attributes} \\

& 
& \bf PSNR $\uparrow$ 
& \bf CD $\downarrow$ 
& \bf F-score $\uparrow$ 
& \bf \textcolor{color2}{Absolute scale} $\downarrow$ 
& \bf \textcolor{color3}{Material} $\uparrow$ 
& \bf \textcolor{color1}{Affordance} $\uparrow$ 
& \bf \textcolor{color4}{Kinematic} $\uparrow$ 
& \bf \textcolor{color5}{Description} $\uparrow$ \\
\midrule

\multirow{5}{*}{\textbf{\ournewdata}} 
& Articulate-Anything~\cite{le2024articulate} & 14.03 & 48.77 & 46.44 & -- & -- & -- &  0.2952  & --\\
& MonoArt~\cite{li2026monoart}               & 19.68 & 7.03 & 85.27 & -- & -- & -- &  0.3805 & --\\
& PhysXGen~\cite{cao2025physx}              & 19.41 & 15.19 & 83.56 & 309.31 & 16.51 & 9.40 & 0.3494 & 11.84\\
& PhysX-Anything~\cite{cao2025physxanything}& 15.97 & 37.06 & 40.46 & 298.19 & 15.65 & 10.50 & 0.4191 & 21.38 \\
& \bf \ourname\ (Ours)                & \textbf{21.52} & \textbf{2.95} & \textbf{91.28} & \textbf{2.79} & \textbf{27.23} & \textbf{21.47} &  \textbf{0.9185} & \textbf{31.05} \\

\midrule

\multirow{5}{*}{\textbf{PhysX-Mobility}} 
& Articulate-Anything~\cite{le2024articulate} & 15.02 & 16.09 & 66.95 & -- & -- & -- & 0.6396 & --\\
& MonoArt~\cite{li2026monoart}              & 16.46 & 6.35 & 87.41 & -- & -- & -- &  0.4351 & --\\
& PhysXGen~\cite{cao2025physx}             & 15.75 & 35.32 & 79.62 & 46.58 & 16.02 & 8.73 & 0.3884 & 11.60 \\
& PhysX-Anything~\cite{cao2025physxanything}& 16.57 & 23.13 & \textbf{89.51} & 22.58 & 22.58 & 16.29 & 0.7852& 26.28 \\
& \bf \ourname\ (Ours)                     & \textbf{18.38} & \textbf{4.70} & 88.50 & \textbf{2.78} & \textbf{24.09} & \textbf{16.58} & \textbf{0.8603}  & \textbf{28.40}\\
\bottomrule
\end{tabular}%
}

\end{table*}

\begin{table*}[t]
\caption{\label{table:ourbench}\textbf{Quantitative comparison with other methods on \ourbench.} The results validate the strong generalization capability of our method on both real-world and complex synthetic images, achieving significant improvements over all competing methods.}
\vspace{-10pt}
\scriptsize
\centering
\resizebox{1\textwidth}{!}{%

\begin{tabular}{@{}c|ccc|ccccc|@{}}
\toprule
 \multirow{2}{*}{\bf Methods} 
& \multicolumn{3}{c|}{\bf Geometry} 
& \multicolumn{5}{c|}{\bf Physical Attributes} \\

& \bf CLIP $\uparrow$ 
& \bf 3D Consistency $\uparrow$ 
& \bf Visual Quality $\uparrow$ 
& \bf \textcolor{color2}{Absolute scale} $\uparrow$ 
& \bf \textcolor{color3}{Material} $\uparrow$ 
& \bf \textcolor{color1}{Affordance} $\uparrow$ 
& \bf \textcolor{color4}{Kinematic} $\uparrow$ 
& \bf \textcolor{color5}{Description} $\uparrow$ \\
\midrule

Articulate-Anything~\cite{le2024articulate} & 0.554 & 55.27& 88.46& -- & -- & -- &71.25  & --\\
MonoArt~\cite{li2026monoart}               & \bf 0.835 &\bf 82.56 &\bf 96.20& -- & -- & -- & 68.32 & --\\
PhysXGen~\cite{cao2025physx}               & 0.803 & 73.50&85.93 & 24.21 & -- & 66.07& 69.17 & 22.24\\
PhysX-Anything~\cite{cao2025physxanything}  & 0.547 &52.71 & 70.81 & 50.20 & 44.70 & 59.96 & 65.99&26.89\\

\bf \ourname\ (Ours)                   & 0.767 & 64.48& 90.0 & \bf 64.26 & \bf 59.89 &\bf 70.57 & \bf80.72&\bf39.02\\

\bottomrule
\end{tabular}%
}
\end{table*}

\begin{figure*}[t]
		\centering	
		
        \includegraphics[width=1\linewidth]{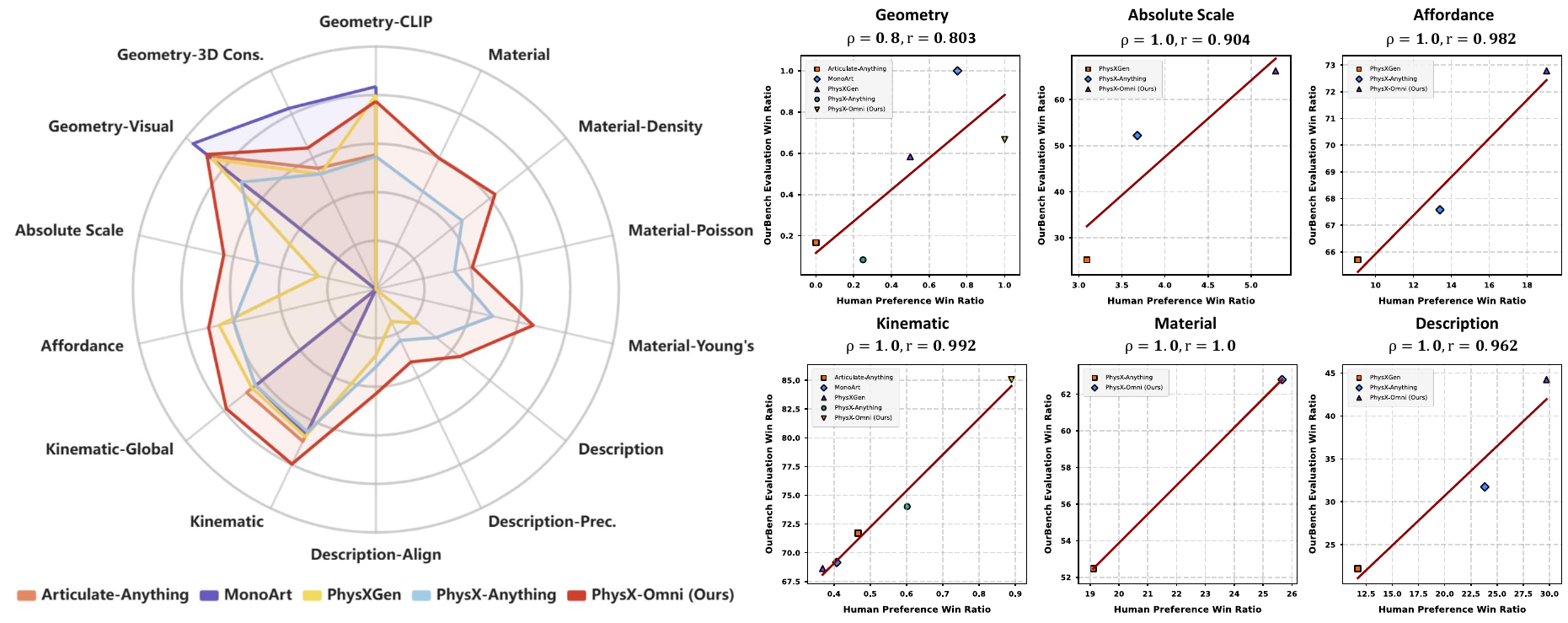}
		\vspace{-8mm}
		\caption{\textbf{Left: Comparison of our \ourname\ with other methods.} It validate the impressive overall performance of \ourname. \textbf{Right: Human alignment validation of \ourbench.} Our experiments show that the our \ourbench\ across all dimensions closely align with human annotations.}
		
		\label{fig:corr}
        \vspace{-10pt}
		
	\end{figure*}

    \begin{figure*}[t]
		\centering	
		
        \includegraphics[width=1\linewidth]{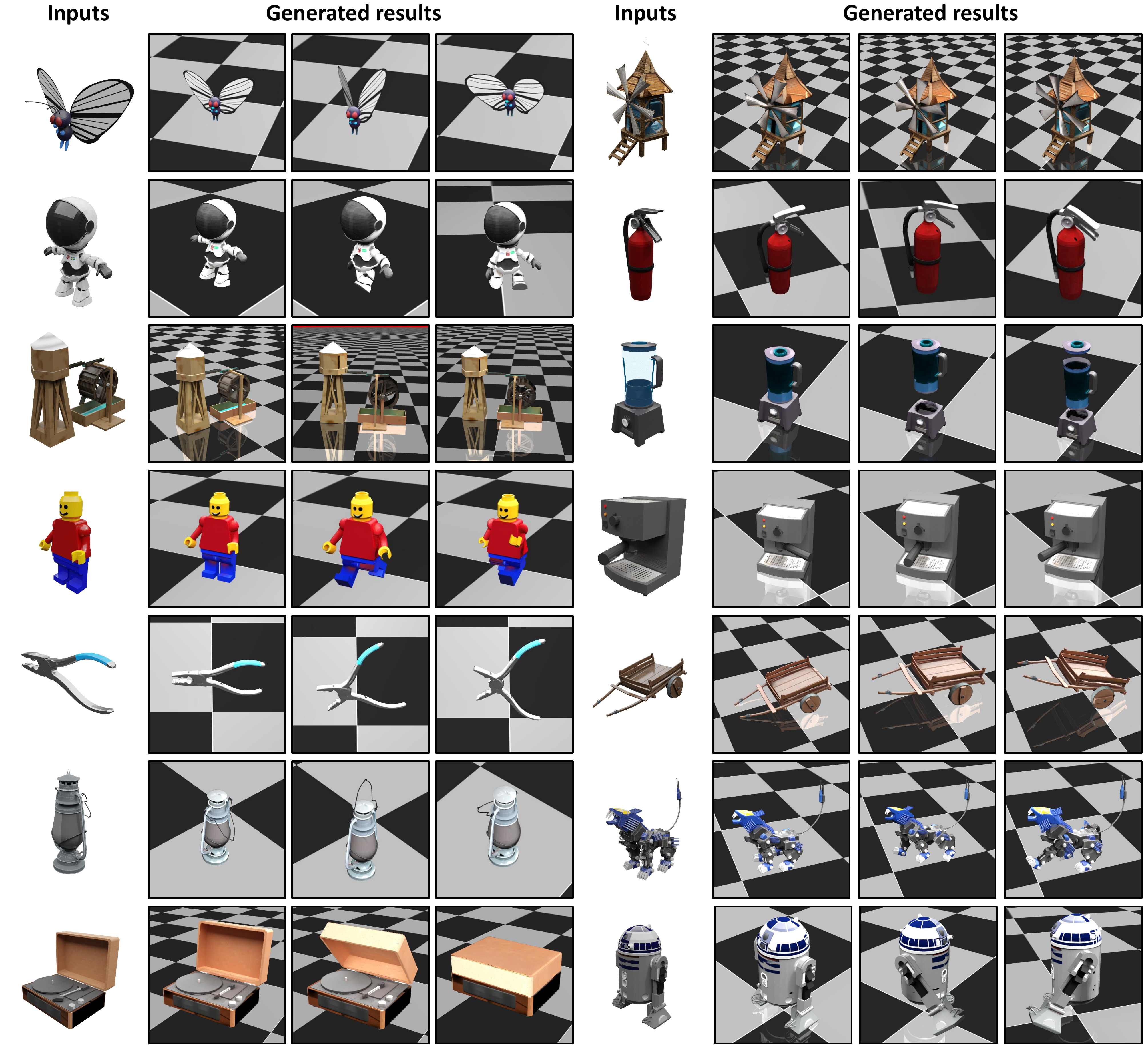}
		\vspace{-6mm}
		\caption{\textbf{More qualitative results of \ourname.} Additional results further demonstrate the robust generative performance of our method in complex scenarios.}
		\vspace{-10pt}
		\label{fig:addquali}

	\end{figure*}
  
    \begin{figure}[t]
		\centering	
		
        \includegraphics[width=0.7\linewidth]{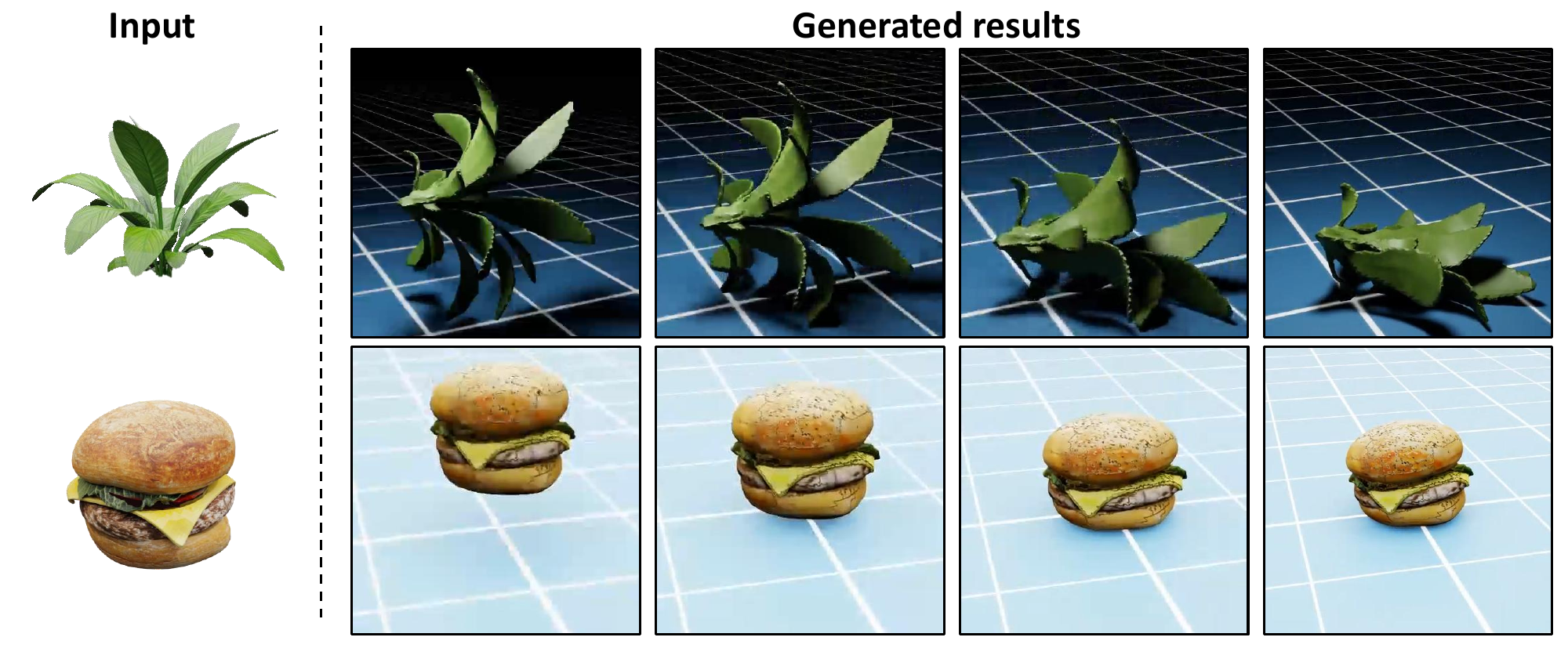}

		\caption{\textbf{Visualization of the generated deformable objects.} It illustrates the realistic deformation behavior of our generated deformable assets during free fall under physical simulation.}
		
		\label{fig:deform}
	\end{figure}

\subsection{Conventional evaluation metrics}
In our experiments, we evaluate the generated simulation-ready 3D assets using both conventional geometric metrics and physical attribute metrics to comprehensively assess visual fidelity, structural quality, and physical correctness. For geometry evaluation, we adopt Peak Signal-to-Noise Ratio (PSNR) to measure rendered appearance quality, and Chamfer Distance (CD) together with F-score to evaluate the accuracy of reconstructed 3D geometry. To ensure robustness and reduce viewpoint bias, we render both the generated assets and the ground-truth assets from 30 different viewpoints and compute the averaged evaluation results across all rendered views. Beyond geometric quality, we further evaluate the generated physical attributes following the protocol of~\cite{cao2025physxanything}. Specifically, for absolute scale evaluation, we compute the Mean Squared Error (MSE) between the predicted and ground-truth object scales. For material, affordance, and description evaluation, we adopt heatmap-based PSNR metrics to measure the similarity between the predicted physical attribute distributions and the corresponding ground-truth annotations. These metrics provide a more robust evaluation of semantic and physical consistency in generated assets. For kinematic evaluation, we measure the MSE between the predicted and ground-truth articulation parameters, including joint axis positions, joint directions, joint types, and motion limits. This evaluation particularly assesses whether the generated assets can accurately capture physically plausible articulated behaviors required for downstream simulation and robotic interaction. By jointly evaluating geometry, physical attributes, and articulation properties, our evaluation protocol provides a comprehensive assessment of simulation-ready physical 3D generation quality.

\subsection{Evaluations with conventional metrics}

We compare \ourname\ with several recent simulation-ready 3D generation methods, including PhysXGen~\cite{cao2025physx}, Articulate-Anything~\cite{le2024articulate}, MonoArt~\cite{li2026monoart}, and PhysX-Anything~\cite{cao2025physxanything}. Following the evaluation protocols of PhysXGen~\cite{cao2025physx} and MonoArt~\cite{li2026monoart}, we conduct experiments on both \ournewdata\ and PhysX-Mobility datasets using conventional geometric metrics and physical attribute evaluations.

As shown in Table.~\ref{table:tradition}, \ourname\ consistently achieves the best performance across nearly all evaluation metrics on both datasets, demonstrating the effectiveness of our unified simulation-ready generation framework. In terms of geometric quality, \ourname\ significantly outperforms previous methods on PSNR, Chamfer Distance (CD), and F-score. On \ournewdata, our method achieves a PSNR of 21.52, CD of 2.95, and F-score of 91.28, substantially surpassing the previous best results. These improvements indicate that our method can generate more accurate and structurally consistent 3D geometry with finer local details. The strong geometric performance mainly benefits from our tailored template-based geometry representation. Unlike previous methods that rely on text-based voxel indices or additional segmentation modules, our representation directly models high-resolution structures in an explicit and compact manner. This design effectively reduces segmentation-induced artifacts and improves the consistency between neighboring geometric regions. As a result, \ourname\ generates cleaner object boundaries, more detailed local structures, and more coherent articulated components, especially for objects with complex topologies and fine-grained geometry.

Beyond geometric generation, \ourname\ also demonstrates substantial improvements in physical attribute prediction. In particular, our method achieves remarkably lower absolute scale errors compared with previous approaches. On \ournewdata, the absolute scale error is reduced from 309.31 in PhysXGen and 298.19 in PhysX-Anything to only 2.79 in \ourname. A similar improvement is observed on PhysX-Mobility, where the error decreases to 2.78. These results demonstrate that our framework possesses significantly stronger understanding of real-world object dimensions and physical priors, which is essential for downstream simulation and robotic interaction tasks. For material, affordance, and description evaluations, \ourname\ also consistently achieves the best performance across both datasets. On \ournewdata, our method improves the material score from 15.65 to 27.23 and the affordance score from 10.50 to 21.47 compared with PhysX-Anything. Similar trends can be observed on PhysX-Mobility. These improvements indicate that \ourname\ can better capture semantic functionality and physical properties of objects, producing more realistic and physically plausible simulation-ready assets. Notably, the most significant gains are achieved in kinematic evaluation. On \ournewdata, \ourname\ reaches a kinematic score of 0.9185, greatly outperforming previous methods such as PhysX-Anything (0.4191) and MonoArt (0.3805). On PhysX-Mobility, our method similarly achieves a strong kinematic score of 0.8603. These results validate that our framework can accurately infer articulation structures, joint types, and motion constraints, enabling the generation of articulated assets with physically consistent behaviors.

Overall, the quantitative results demonstrate that \ourname\ achieves superior performance in both geometry and physical reasoning. Benefiting from the combination of our VLM-based global-to-local framework and the proposed high-resolution geometry representation, \ourname\ produces simulation-ready assets with higher visual fidelity, stronger physical consistency, and more accurate articulation modeling. These results collectively validate the superiority, robustness, and generalizability of our unified framework for simulation-ready physical 3D generation across diverse object categories and physical settings.

\subsection{Evaluations on \ourbench}

To comprehensively evaluate the generalization ability of different methods in real-world scenarios, we further conduct evaluations on our newly proposed benchmark, \ourbench. More details of the benchmark construction and evaluation metrics are provided in the supplementary material. As mentioned previously, the conditioning images in \ourbench\ are collected from both real-world photographs and rendered images of 3D assets, covering a wide range of common object categories and challenging in-the-wild cases. Compared with conventional benchmarks that rely on ground-truth annotations, \ourbench\ emphasizes ground-truth-free evaluation across multiple dimensions, including geometry, absolute scale, material, affordance, kinematics, and semantic description.

The quantitative results in Table~\ref{table:ourbench} strongly demonstrate the superior overall performance of \ourname\ compared with existing approaches. In particular, our method achieves the best results on most physical attributes, including absolute scale, material, affordance, kinematics, and description. Specifically, \ourname\ achieves a kinematic score of 80.72, significantly outperforming PhysX-Anything (65.99), PhysXGen (69.17), MonoArt (68.32), and Articulate-Anything (71.25). Similar improvements can also be observed for affordance understanding and description quality, validating the strong physical reasoning and semantic understanding capabilities of our framework.

Benefiting from the explicitly encoded high-resolution 3D structures, \ourname\ can better model the intrinsic interdependency between geometry, articulation, and physical attributes. Unlike prior methods that heavily rely on segmentation-based intermediate representations, our framework directly models explicit 3D structures in a unified manner, thereby significantly improving structural coherence and articulation consistency. This advantage is particularly important for complex articulated and deformable objects, where geometric details and kinematic properties are highly coupled. Although MonoArt achieves slightly better performance on several geometry-related metrics, including CLIP similarity, 3D consistency, and visual quality, this advantage mainly arises from its complete reliance on the pre-trained TRELLIS geometry generation pipeline. As a result, MonoArt lacks explicit understanding of part-level motion and physical interactions. Consequently, it exhibits limited capability in modeling physical properties and articulated behaviors, leading to notably inferior performance on simulation-oriented attributes such as kinematics, affordance, and absolute scale. In comparison, \ourname\ achieves a much more balanced and robust performance across both geometry and physical reasoning tasks shown in Fig.~\ref{fig:corr}. The results demonstrate that our method not only preserves strong geometric quality, but also substantially improves the generation of physically plausible simulation-ready assets. In particular, our explicit geometry representation enables the model to maintain detailed structures while avoiding segmentation-induced ambiguities and artifacts, resulting in more coherent articulated motions and more reliable physical attributes.

To provide more intuitive comparisons, we further visualize the generated results of different methods in Fig.~\ref{fig:quali}. The qualitative results show that \ourname\ demonstrates remarkable robustness in generating complex structures and challenging articulated objects. Compared with the baseline PhysX-Anything, our method produces higher-quality simulation-ready assets with more accurate geometry, more plausible material and affordance predictions, and significantly more coherent articulated behaviors. Furthermore, additional visualization results in Fig.~\ref{fig:addquali} and Fig.~\ref{fig:deform} demonstrate the capability of \ourname\ in generating diverse simulation-ready scenes involving rigid, deformable, and articulated objects. By directly modeling explicit 3D structures without relying on additional segmentation modules, \ourname\ effectively reduces structural ambiguities and segmentation-induced errors, further validating the effectiveness of our tailored geometry representation and unified simulation-ready generation framework.

\begin{figure}[t]
		\centering	
		
        \includegraphics[width=0.5\linewidth]{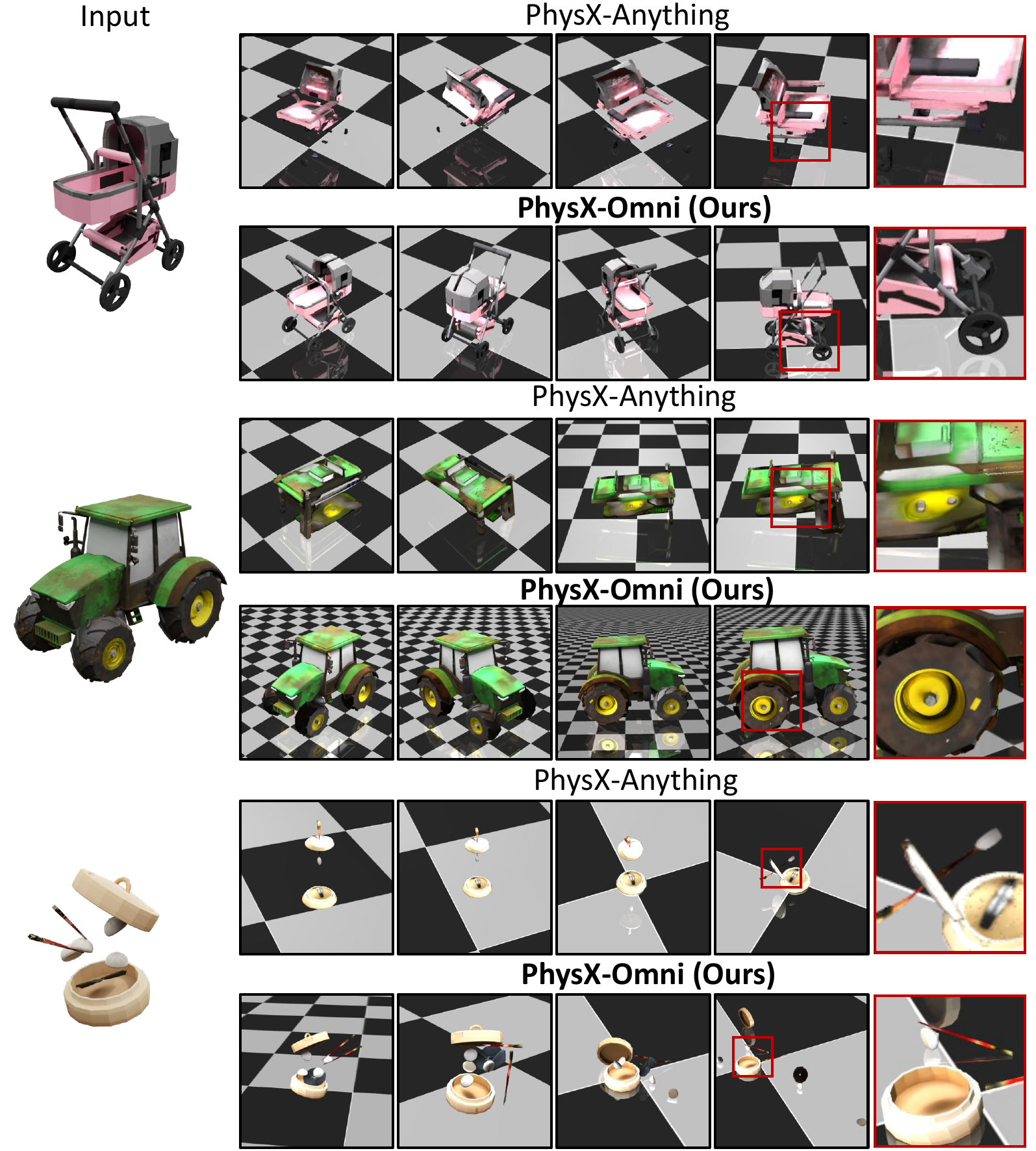}
		\vspace{-2mm}
		\caption{\textbf{Visualization of model using different geometry representations.} It strongly demonstrates that, by introducing an efficient geometry representation, \ourname\ achieves superior performance in generating complex structures compared with our baseline.}
		
		\label{fig:ablation}

	\end{figure}

\begin{figure*}[t]
		\centering	
		
        \includegraphics[width=1\linewidth]{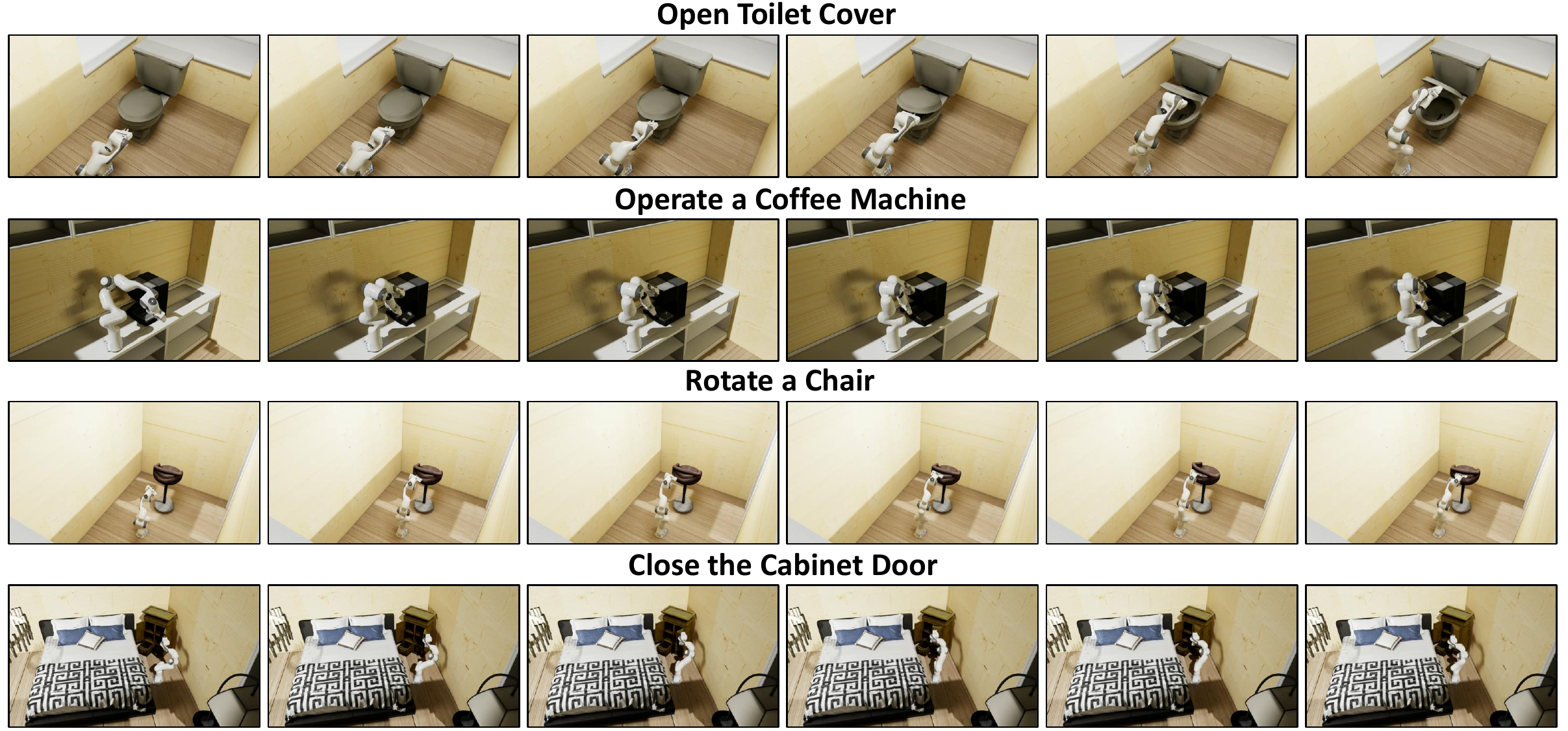}
		\vspace{-6mm}
		\caption{\textbf{Robot Manipulation on our Generated sim-ready 3D assets.} The results demonstrate that our generated simulation-ready assets exhibit highly physically plausible behaviors and accurate geometric structures across diverse tasks, thereby opening a new direction for robotic policy learning.}
		
		\label{fig:sim}
 
	\end{figure*}
    \begin{figure*}[t]
		\centering	
		
        \includegraphics[width=1\linewidth]{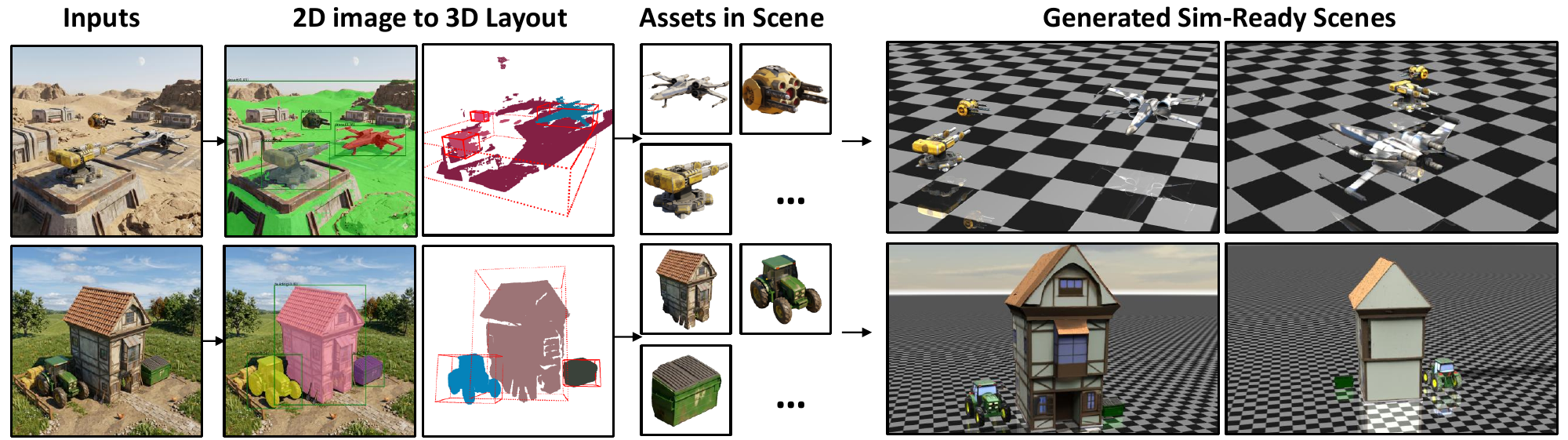}
		\vspace{-7mm}
		\caption{\textbf{Applications of our \ourname.} We explore the potential applications of \ourname\ in sim-ready scene generation.}
		\vspace{-10pt}
		\label{fig:scene}
		
	\end{figure*}
\subsection{Validating human alignment of \ourbench} 

To validate that \ourbench\ can effectively reflect human perception and evaluation preferences, we further study the correlation between the automatic evaluation results produced by \ourbench\ and human annotations. Specifically, following prior benchmark protocols, we measure the alignment between automatic evaluation scores and human preference scores using Spearman’s rank correlation coefficient. A higher Spearman correlation coefficient indicates stronger consistency between the benchmark evaluation and human judgment. As shown in Fig.~\ref{fig:corr}, \ourbench\ demonstrates consistently strong correlations with human annotations across all major evaluation dimensions, including geometry, absolute scale, affordance, kinematics, material, and semantic description. In particular, several physical attributes achieve superior rank consistency with human preferences. For example, absolute scale, affordance, material, and description all achieve a Spearman coefficient of $\rho=1.0$, while kinematic evaluation reaches $\rho=1.0$ with an exceptionally high Pearson correlation coefficient of $r=0.992$. These results indicate that the rankings produced by \ourbench\ are highly aligned with human evaluation outcomes. Moreover, even for geometry evaluation, which is generally more challenging due to the diversity of visual appearances and structural details, \ourbench\ still achieves a strong correlation with human preferences ($\rho=0.8$, $r=0.803$). This result further demonstrates that our benchmark can robustly evaluate not only physical attributes but also geometric quality in complex real-world scenarios.

Overall, the strong correlations across all evaluation dimensions validate the reliability, robustness, and effectiveness of \ourbench. These results demonstrate that our benchmark can serve as a trustworthy automatic evaluation framework for simulation-ready physical 3D generation, providing evaluation results that closely match human perception and judgment.

\subsection{Ablation Studies}

To validate the effectiveness of our tailored geometry representation, we compare \ourname\ with a baseline that directly employs text-based voxel indices to model 3D structures. The quantitative results on both conventional metrics and \ourbench, reported in Table~\ref{table:tradition} and Table~\ref{table:ourbench}, consistently demonstrate the substantial improvements brought by our proposed template-based geometry representation. In particular, our method achieves significantly better performance on kinematic and absolute scale, validating the effectiveness of explicitly modeling high-resolution structures in a compact and generation-friendly manner.

Furthermore, the qualitative comparisons shown in Fig.~\ref{fig:ablation} provide more intuitive evidence of the advantages of our representation. Compared with the baseline PhysX-Anything, which relies on text-based voxel indices and additional segmentation processes, \ourname\ produces substantially more detailed and structurally coherent simulation-ready assets. As illustrated in the highlighted regions, the baseline method frequently suffers from structural ambiguities, incomplete local geometry, and inconsistent articulated components, especially for objects with complex topologies and fine-grained structures, such as strollers and tractors. In contrast, by directly modeling explicit 3D geometry and eliminating the additional segmentation stage during generation, \ourname\ effectively reduces segmentation-induced artifacts and error accumulation. This design enables our method to better preserve local geometric continuity and part-level structural consistency, leading to sharper details, more accurate articulated structures, and more physically plausible object layouts. For example, \ourname\ can generate more accurate wheel structures, cleaner articulated connections, and more stable local geometries in highly complex regions where the baseline often fails.

Moreover, the improvements are particularly evident for articulated objects and objects involving strong part interactions. Benefiting from the explicit structural representation, \ourname\ can better capture the intrinsic relationships between geometry and kinematics, thereby improving both structural reasoning and motion consistency. These results collectively demonstrate that our tailored geometry representation significantly enhances the robustness, fidelity, and generalization ability of simulation-ready physical 3D generation.

\subsection{Application: Robotic Policy Learning in Simulation}

As shown in Fig.~\ref{fig:sim}, we further investigate whether the generated assets can be effectively utilized in real simulation environments and downstream robotic tasks. To this end, we directly deploy the generated simulation-ready 3D assets into a physics simulator for robotic interaction and policy learning. Specifically, the generated assets are imported together with their geometric structures, physical properties, and articulated parameters, enabling the simulator to perform physically grounded interactions without additional manual processing. The experimental results  demonstrate that our generated assets maintain reliable geometric accuracy, physically plausible material properties, and coherent articulated behaviors under dynamic interactions. Even in challenging manipulation scenarios involving articulated motion and object contact, the generated assets remain structurally stable and physically consistent. These results suggest that \ourname\ not only produces visually realistic 3D assets, but also generates physically functional representations that can be seamlessly integrated into simulation pipelines for downstream robotics and embodied AI applications. Furthermore, the ability to automatically generate simulation-ready assets from in-the-wild images significantly reduces the cost of manual asset construction for robotic training environments.

\subsection{Application: Sim-Ready Scene Generation}

In addition, we further explore the potential of \ourname\ for scene-level simulation-ready generation. Specifically, we first employ image-to-depth estimation methods~\cite{yang2024depth} together with 2D segmentation approaches~\cite{ravi2024sam} to reconstruct an initial 3D scene layout from input images. Based on the estimated depth, segmentation masks, and scene geometry, we obtain coarse object placements and spatial relationships within the environment. We then integrate the reconstructed 3D layout with the simulation-ready assets generated by \ourname\ to automatically build physically plausible simulation-ready scenes, as illustrated in Fig.~\ref{fig:scene}. Benefiting from the explicit geometric structures and physical attributes generated by our framework, the inserted assets can maintain consistent scales. Moreover, since our framework supports rigid, deformable, and articulated objects in a unified manner, it enables the construction of significantly more diverse and realistic simulation environments compared with previous approaches.

These results demonstrate that \ourname\ not only supports high-quality simulation-ready asset generation, but also provides a promising foundation for scalable scene-level simulation construction, embodied AI training, robotic policy learning, and future physically grounded world generation applications.

\section{Conclusion}

In this paper, we introduce \textbf{\ourname}, a unified framework for simulation-ready physical 3D generation across diverse asset types, including rigid, deformable, and articulated objects. By proposing a tailored geometry representation for vision--language models, \ourname\ directly models detailed 3D structures without introducing additional special tokens or relying on segmentation modules, thereby significantly improving generation quality and robustness. To alleviate the limitation of data scarcity, we further construct the first general simulation-ready physical 3D dataset, \ournewdata, containing over 8.7K high-quality assets with rich physical annotations. Moreover, to evaluate simulation-ready 3D generation in real-world scenarios, we propose a new benchmark, \ourbench, which performs ground-truth-free evaluation across six key dimensions, including geometry, absolute scale, affordance, material, kinematics, and semantic description. Comprehensive experiments on both \ourbench\ and conventional evaluation metrics demonstrate the superior performance and strong generalization ability of \ourname. Furthermore, additional studies validate the potential of our framework in downstream applications such as simulation-ready scene generation and robotic policy learning, highlighting the feasibility of directly deploying generated assets into embodied AI and robotic simulation environments.

\noindent\textbf{Limitation and future work.} Despite the strong performance of \ourname, several limitations remain. In particular, geometric quality can still be improved for highly complex structures and fine-grained details. Since our framework emphasizes unified physical understanding and simulation-ready generation rather than appearance-oriented geometry pre-training, it may underperform on certain appearance-focused geometric metrics. In the future, we plan to leverage larger-scale 3D geometry datasets and stronger appearance supervision to further enhance geometric fidelity while maintaining physical consistency.

\subsection{Acknowledgments}
This research is supported by cash and in-kind funding from NTU S-Lab and industry partner(s). This study is also supported by the Ministry of Education, Singapore, under its MOE AcRF Tier 2 (MOE-T2EP20221-0012, MOE-T2EP20223-0002). 

\bibliographystyle{unsrtnat}
\bibliography{reference}

\end{document}